\documentclass[letterpaper, 10 pt, conference,table,xcdraw]{ieeeconf}  

\IEEEoverridecommandlockouts                              

\usepackage{authblk}
\usepackage{graphics} 
\usepackage{epsfig} 
\usepackage{mathptmx} 
\usepackage{times} 
\usepackage{amsmath} 
\usepackage{amssymb}  
\usepackage{cite}
\usepackage{booktabs}
\usepackage{float}
\usepackage{multirow}
\usepackage{caption}
\usepackage{newtxtext, newtxmath}
\usepackage{tabularx}
\usepackage{longtable}
\usepackage{hyperref}

\usepackage{flushend}
\usepackage{stfloats}
\usepackage{balance}
\usepackage{xspace}
\usepackage{listings}
\usepackage{algorithm}
\usepackage{algpseudocode}
\usepackage{tcolorbox}
\usepackage{graphicx}

\title{\LARGE \bf
RoboEngine: Plug-and-Play Robot Data Augmentation with Semantic Robot Segmentation and Background Generation
}

\author{Chengbo Yuan*$^{1,3}$, Suraj Joshi*$^{2}$, Shaoting Zhu*$^{1,3}$, Hang Su$^{2}$, Hang Zhao$^{1,3}$, Yang Gao†$^{1,3}$\\
\href{https://roboengine.github.io/}{\large\textbf{RoboEngine.github.io/}}
\thanks{$^{1}$ Institute for Interdisciplinary Information Sciences, Tsinghua University}%
\thanks{$^{2}$ Department of Computer Science and Technology, Tsinghua University}%
\thanks{$^{3}$ Shanghai Qi Zhi Institute}%
\thanks{* These authors contributed equally to this work.}
\thanks{† Corresponding at: \texttt{gaoyangiiis@mail.tsinghua.edu.cn}}%
}

\begin{document}

\newcommand{\insertteaser}{
    \includegraphics[width=\linewidth]{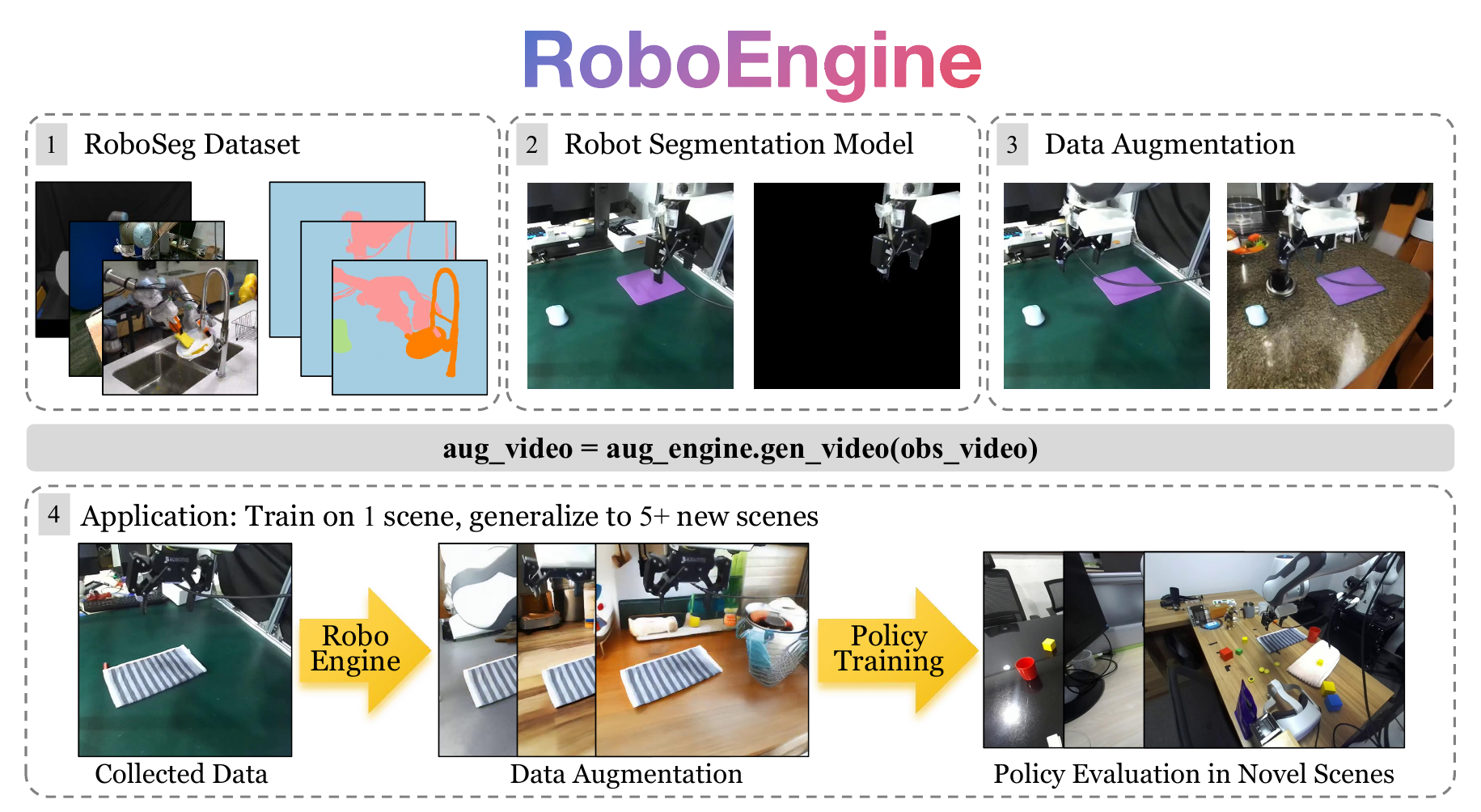}
    \captionof{figure}{We introduce \texttt{RoboEngine}, the first plug-and-play toolkit for visual robot data augmentation, enabling physics- and task-aware robot scene generation without any prerequisites. To achieve this, we propose novel datasets, models, and an out-of-the-box toolkit. Extensive real-robot experiments validate the effectiveness of \texttt{RoboEngine}.}
    \label{fig: method-overview}
}

\makeatletter
\apptocmd{\@maketitle}{\centering\insertteaser}{}{}
\makeatother

\maketitle
\setcounter{figure}{1}

\thispagestyle{empty}
\pagestyle{empty}



\begin{abstract}
Visual augmentation has become a crucial technique for enhancing the visual robustness of imitation learning. However, existing methods are often limited by prerequisites such as camera calibration or the need for controlled environments (e.g., green screen setups). In this work, we introduce \texttt{RoboEngine}, the first plug-and-play visual robot data augmentation toolkit. For the first time, users can effortlessly generate physics- and task-aware robot scenes with just a few lines of code. To achieve this, we present a novel robot scene segmentation dataset, a generalizable high-quality robot segmentation model, and a fine-tuned background generation model, which together form the core components of the out-of-the-box toolkit. Using \texttt{RoboEngine}, we demonstrate the ability to generalize robot manipulation tasks across six entirely new scenes, based solely on demonstrations collected from a single scene, achieving a more than 200\% performance improvement compared to the no-augmentation baseline. All datasets, model weights, and the toolkit are released \href{https://roboengine.github.io/}{https://roboengine.github.io/}.

\end{abstract}

\section{INTRODUCTION}

Visuomotor robot manipulation systems have advanced rapidly in recent years~\cite{visuomotor}. One of the key methods driving this progress is imitation learning~\cite{aloha, dp}. However, imitation learning is widely known to suffer from visual disturbances, particularly when the training environment is limited~\cite{decompose, greenaug}. While collecting large-scale datasets across diverse scenes can alleviate this issue~\cite{open_x_embodiment_rt_x_2023, droid, data_scaling_law}, this approach is not scalable due to the high costs in terms of time and resources. On the other hand, data augmentation methods~\cite{greenaug, semantic_aug, rosie}, particularly generative augmentation techniques, offer a more scalable solution by effortlessly generating diverse visual backgrounds for model training, and have thus gained significant popularity.

Despite their efficiency, existing robot data augmentation methods do not support \textbf{plug-and-play usage without pre-requisite limitations} (e.g., like the ColorJitter function for computer vision tasks), which restricts their adoption and dissemination within the community. This limitation arises from a lack of \textbf{simultaneous convenience, realism, and diversity}. First, many methods impose pre-requisite conditions for foreground mask generation. For example, GreenAug~\cite{greenaug} requires a green screen, and CACTI~\cite{semantic_aug} demands camera calibration. Furthermore, many inpainting-based methods~\cite{semantic_aug, roboagent} are limited to small object-level augmentations and cannot alter the entire background or layout. Some methods even directly modify the scene using random images or texture~\cite{greenaug}, which fail to respect physical constraints, leading to degenerated real-world performance due to distribution shifts~\cite{scaling_vs}.

To address these limitations, we propose \texttt{RoboEngine}, \textbf{the first plug-and-play toolkit for physics- and task-aware robot visual data augmentation}. With \texttt{RoboEngine}, users can generate physics-constrained backgrounds for robot data \textbf{with just a few lines of code}, and without any pre-requisite conditions. We identify that the lack of plug-and-play robot mask generation is a major bottleneck in existing methods~\cite{rovi-aug}. To solve this, we introduce \texttt{RoboSeg}, a novel segmentation dataset consisting of 3,800 high-quality annotated robot scene images. By fine-tuning a state-of-the-art (SoTA) model~\cite{evf} on this dataset, we create the first generalizable ``wire-level'' robot segmentation model. Additionally, we fine-tune a Diffusion Model~\cite{bg_diffusion} on our dataset to enable task-aware background generation. These components are encapsulated into a convenient plug-and-play toolkit, similar to the ColorJitter function in the computer vision field\footnote{pytorch.org/vision/main/generated/torchvision.transforms.ColorJitter.html}.

To validate the effectiveness of \texttt{RoboEngine}, we integrate several SoTA methods~\cite{greenaug, semantic_aug, dp} into our plug-and-play toolkit, removing all pre-requisite limitations, and conduct real-robot experiments. Training on human demonstrations collected \textbf{from only a single scene}, our method enables generalizable manipulation across six entirely new scenes, with significant changes in background, layout, lighting, etc. This results in a 210\% and 20\% performance improvement over the no-augmentation baseline and the previous SoTA method (our plug-and-play implementation). Additionally, we observe that as the number of augmented demonstrations increases, performance continues to improve to some extent.

In summary, we aim to provide a plug-and-play augmentation toolkit for the robot learning community, akin to the role of ColorJitter in computer vision. Our main contributions are as follows:

\begin{itemize}
    \item We introduce a novel dataset, \texttt{RoboSeg}, which contains 3,800 high-quality segmentation images of robot scenes. Based on this dataset, we train the first generalizable high-quality robot segmentation model. 
    \item We integrate our segmentation model and a new fine-tuned physics- and task-aware Diffusion model into the plug-and-play \texttt{RoboEngine} toolkit. Users can achieve data augmentation with just a few lines of code, without any pre-requisite limitations.
    \item We conduct extensive real-robot experiments to validate the effectiveness of \texttt{RoboEngine}. Training on demonstrations from only one scene, our method enables generalizable manipulation across six entirely new scenes, achieving significant improvements over previous methods. 
\end{itemize}

\section{RELATED WORKS}

\subsection{Visual Generalizable Imitation Learning}

Imitation Learning (IL) \cite{aloha, open-television, rt2, openvla} has become a key approach for visuomotor robot control \cite{visuomotor, deep_visuomotor}, where deep neural networks are used to predict motor actions directly from visual inputs. In this paper, we adopt Diffusion Policy (DP) \cite{dp}, a state-of-the-art IL method, as our robot policy architecture. However, raw DP is sensitive to visual disturbances \cite{decompose, greenaug}. To improve visual robustness and generalization in IL, several strategies have been proposed, including novel architectures \cite{dp3, idp3, rise}, vision pretraining \cite{r3m, rpr}, and enhanced data collection and synthesis approaches \cite{data_scaling_law, semantic_aug, cyberdemo}. Our work focuses on addressing these challenges through data augmentation \cite{greenaug} using a plug-and-play robot data synthesis engine.

\subsection{Robot Data Augmentation}

Collecting robot data is both costly and labor-intensive \cite{oxe}, especially in diverse environments \cite{droid}. To mitigate this, various data augmentation techniques \cite{greenaug, semantic_aug, roboagent, robosaga, rosie, cyberdemo, pa_aug, 3d_aug, rovi-aug} have been proposed, enabling the synthesis of diverse robot data at minimal cost. Traditional methods, such as Cropping and ColorJitter, are effective for in-domain visual generalization \cite{dp}, but struggle when handling large visual changes. Generative methods \cite{greenaug} can better manage such changes, but they are not plug-and-play due to their pre-requisite requirements \cite{cyberdemo}. For example, \cite{semantic_aug, roboagent} necessitates camera calibration. It also depends on scene objects for inpainting, which can lead to unpredictable behavior unless hyperparameters are carefully adjusted for each scene. Additionally, these methods \cite{semantic_aug, roboagent, rosie, robosaga} offer limited visual variation, which may fail under extreme conditions. While GreenAug \cite{greenaug} can generate more significant visual changes, it requires a green screen setup and struggles in unstructured environments \cite{droid}. Moreover, it neglects the physical reliability of the environment, which can lead to out-of-distribution issues \cite{scaling_vs}.

\section{METHOD}

In this section, we provide details of our plug-and-play \texttt{RoboEngine} augmentation methods. To generate robot masks without any pre-requisites, we introduce a new robot scene segmentation dataset and develop the first generalizable high-quality robot segmentation model. We then fine-tune a Diffusion Model~\cite{bg_diffusion} and integrate both models into a seamless plug-and-play pipeline.

\subsection{Task Definition}

Given a robot manipulation demonstration dataset $\mathcal{D}={I_1, J_1, I_2, J_2, \dots, I_n, J_n}$, where $I_i$ denotes image data and $J_i$ represents associated information such as language instructions, robot proprioception, and actions, \texttt{RoboEngine} aims to augment $\mathcal{D}$ into $\mathcal{D}_a$ using a trained segmentation model and a generative model.

For each image $I_i = \{R_i, O_i, B_i\}$ in $\mathcal{D}$, where $R_i$ is the robot arm area, $O_i$ is the task-related objects area (e.g., for ``Put Mouse to the Pad,'' $O_i$ includes the mouse and pad), and $B_i$ is the background, we first apply semantic segmentation to split the areas. We then generate a new background $B_i^*$. Theoretically, the closer the distribution of $B_i^*$ is to the robot's deployment environment, the better the robot's manipulation performance~\cite{scaling_vs}. Finally, we combine $R_i$, $O_i$, and $B_i^*$ to create the augmented image $I_i^{*} = \{R_i,O_i,B_i^*\}$. The resulting augmented dataset is represented as $\mathcal{D}_a = \{I_1^*, J_1, I_2^*, J_2, \dots, I_n^*, J_n\}$.

\subsection{RoboSeg Dataset and Robot Segmentation Model}

As noted in previous work~\cite{rovi-aug}, we observe that existing off-the-shelf segmentation models, such as the Segment Anything Model V2 (SAM-V2)\cite{sam2} and LISA\cite{lai2024lisa}, struggle to segment robots with high accuracy. This limitation often necessitates prerequisites such as a green screen setup~\cite{greenaug} or camera calibration~\cite{semantic_aug, roboagent} for generating accurate robot masks.

\begin{figure}[t]
    \centering
    \includegraphics[width=1.0\linewidth]{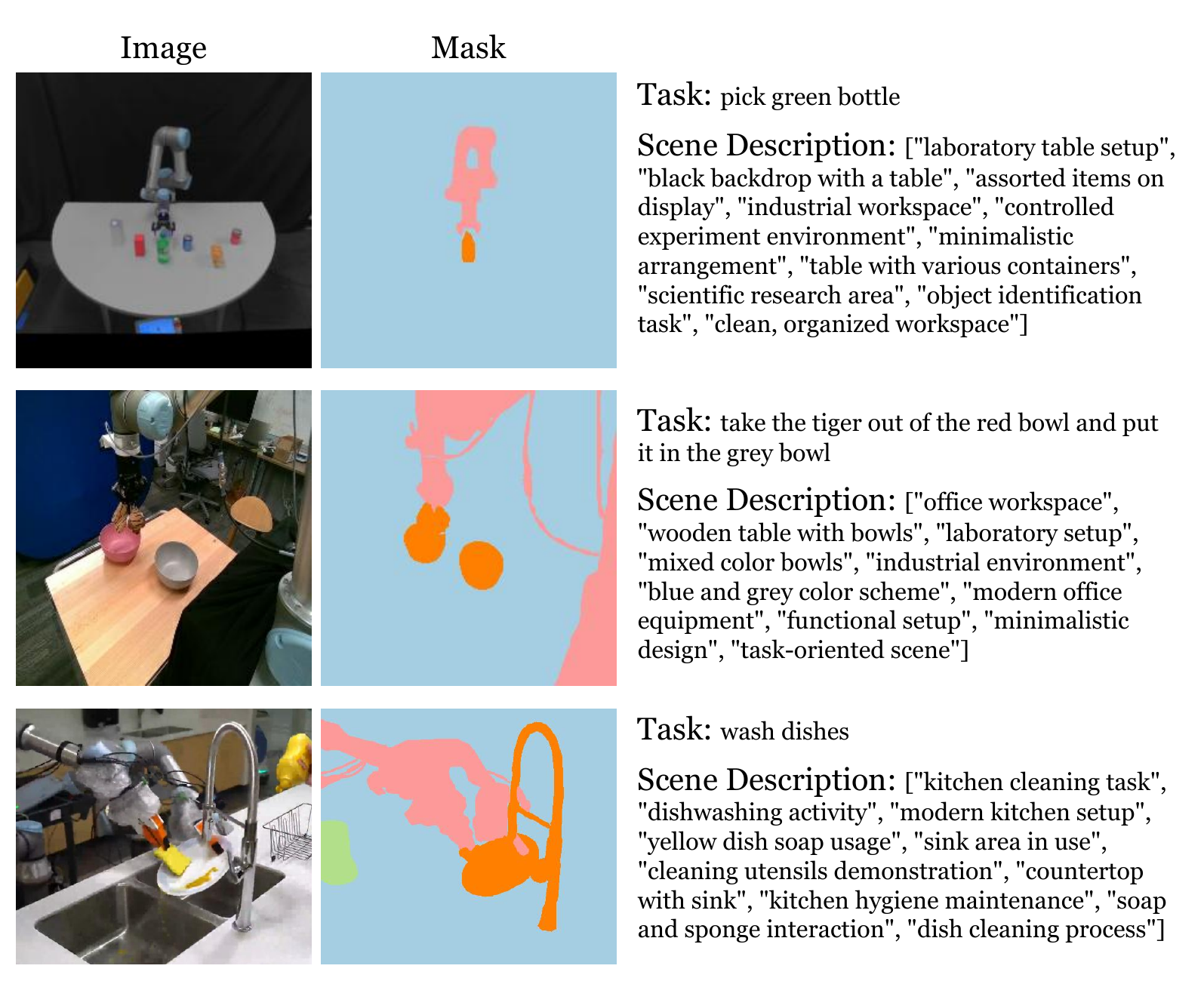}
    \vspace{-5mm}
    \caption{Our \texttt{RoboSeg} dataset provides high-quality, fine-grained semantic segmentation annotations, covering a wide diversity of robots and environments.}
    \vspace{-5mm}
    \label{fig:roboseg}
\end{figure}

To address this, we propose \texttt{RoboSeg}, a novel dataset with high-quality robot scene segmentation annotations. \texttt{RoboSeg} contains 3,800 images randomly selected from over 35 robot datasets \cite{open_x_embodiment_rt_x_2023, droid, jiang2024roboexp, huang2024rekep, rdt, dp}, covering a broad range of robot types (e.g., Franka, WindowX, HelloRobot, UR5, Sawyer, Xarm, etc.), camera views, and background environments. A visualization of \texttt{RoboSeg} can be found in \autoref{fig:roboseg}. For each image, we provide three types of masks: (1) ``robot-main'': the part of the robot arm connected to the robot gripper in pixel space, (2) ``robot-auxiliary'': other parts of the robot, such as the robot base in some images, (3) ``object'': the area of all task-related objects. Note that even the robot’s wires are annotated, creating \textbf{extremely fine-grained masks}. Additionally, each image includes the task instruction, and we use GPT-4o~\cite{gpt-4o} to generate 10 brief descriptions for each scene.

Based on \texttt{RoboSeg}, we fine-tune the state-of-the-art (SoTA) language-conditioned segmentation model EVF-SAM~\cite{evf} to create a new robot segmentation model, \texttt{Robo-SAM}, which achieves high-quality open-world robot segmentation. We choose the language-conditioned version rather than the original SAM~\cite{sam2} because we find that language provides additional visual cues that enhance the base model's performance. Details on model training can be found in Appendix~\ref{app: model_training}.

 \subsection{Plug-and-Play RoboEngine Augmentation}

Given a robot scene image, we first generate the robot mask and task-related object masks using \texttt{Robo-SAM} and EVF-SAM~\cite{evf} (conditioned on the object name in the task instruction). We then use a generative model to create a physics- and task-aware background based on the previously generated masks. For this, we use BackGround-Diffusion~\cite{bg_diffusion}, which generates a foreground-aware background with physical constraints given a foreground mask and scene description. We fine-tune this model on our \texttt{RoboSeg} dataset to eliminate unreasonable generations in some cases. More details can be found in Appendix~\ref{app: model_training}. For scene descriptions, we collect all descriptions from \texttt{RoboSeg} to create a prompt pool, from which we randomly select one during usage to generate a randomly new scenes.

We integrate our segmentation and background generation models into a plug-and-play toolkit, \texttt{RoboEngine}, which requires no pre-requisite setup. Users can generate augmented robot data with just a few lines of code:

\lstset{
    columns=fixed,       
    numbers=left,                
    numberstyle=\tiny\color{gray},   
    numbersep=0.4em,
    frame=none,                      
    backgroundcolor=\color[RGB]{230,230,230},   
    keywordstyle=\color[RGB]{40,40,255},      
    numberstyle=\scriptsize\color{darkgray},          
    commentstyle=\it\color[RGB]{0,96,96},             
    stringstyle=\rmfamily\slshape\color[RGB]{128,0,0},
    showstringspaces=false,
    language=python,       
    basicstyle=\footnotesize\ttfamily, 
    breaklines=true, 
    breakatwhitespace=false 
}
\begin{lstlisting}
from robo_engine import RoboEngine
# aug_method: engine, texture, imagenet......
aug_method = 'robo_engine'
engine = RoboEngine(
    robo_seg_method=['robo_sam_video'],
    obj_seg_method=['evf_sam_video'],
    aug_method=aug_method,
    batch_size=32
)

aug_video = aug_engine.gen_video(obs_video)
\end{lstlisting}

This is similar to the usage of fundamental traditional augmentation methods, such as ColorJitter or RandomCrop, in the computer vision community. We hope this will facilitate the adoption and wider use of generative visual robot data augmentation methods.

\section{EXPERIMENTS}

\subsection{Segmentation Results}

\textbf{Experiment Setting.}\ \ We first evaluate our new robot segmentation model, \texttt{Robo-SAM}, against other off-the-shelf state-of-the-art (SoTA) segmentation models. We evaluate all models on two datasets: (1) ``Test Set'', which contains 97 images of novel scenes extracted from the same source sub-dataset as our proposed \texttt{RoboSeg} dataset, and (2) ``Zero-shot Set'', which consists of 45 robot manipulation images randomly downloaded from the internet. We compare our model with three widely used language-conditioned SoTA segmentation models: ClipSeg~\cite{luddecke2022image}, LISA~\cite{lai2024lisa}, and EVF-SAM~\cite{evf}, all using the ``robot'' instruction prompt. Generalized Intersection over Union (GIoU)~\cite{giou} is used as the evaluation metric for our quantitative experiments.

\begin{figure}[t]
    \centering
    \includegraphics[width=1.0\linewidth]{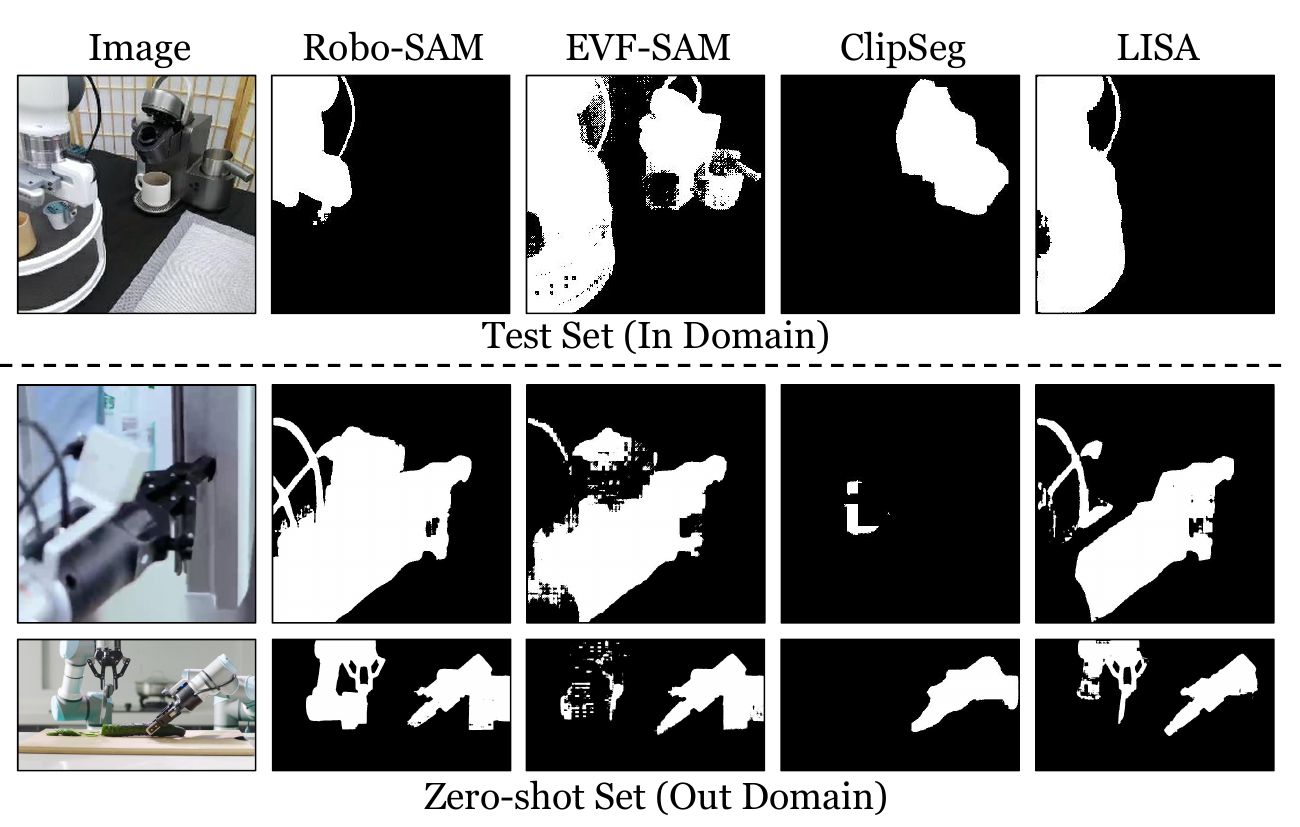}
    \vspace{-2mm}
    \caption{Comparison of Segmentation Results between our new \texttt{Robo-SAM} model and other baselines. Only \texttt{Robo-SAM} produces usable segmentation masks for downstream augmentation applications.}

    \vspace{-3mm}
    \label{fig:seg_example}
\end{figure}

\textbf{Results.}\ \ The quantitative results are shown in \autoref{tab:segmentation}. Our new \texttt{Robo-SAM} model outperforms all other baselines with a substantial improvement ($>$ 0.12 GIoU) on both the ``Test'' and ``Zero-shot'' sets. We also visualize the predictions of all models in \autoref{fig:seg_example}. As shown, none of the baselines achieve usable robot segmentation, as observed in previous work \cite{rovi-aug}. In contrast, our \texttt{Robo-SAM} model aligns well with the robot shape, even in challenging novel scenes. This opens new possibilities for calibration-free and plug-and-play robot segmentation and augmentation techniques.

\begin{table}[!t]
    \vspace{2mm}
    \centering
    \caption{Segmentation Performance Comparison of different methods on the Test Set and Zero-shot Set, evaluated using the Generalized IoU (GIoU) metric~\cite{giou}.}
    \vspace{-1mm}
    \label{tab:segmentation}
    \setlength\tabcolsep{8pt}
    \fontsize{8}{10}\selectfont
    \begin{tabular}{p{2.3cm}<{\centering}|p{2.0cm}<{\centering}|p{2.0cm}<{\centering}}
        \toprule
        \textbf{Method} & \textbf{Test Set} $\uparrow$ & \textbf{Zero-shot Set} $\uparrow$ \\
         \midrule
         CLIPSeg\cite{luddecke2022image} & 0.2810 & 0.4049 \\
         LISA\cite{lai2024lisa} & 0.6040 & 0.7571 \\

         EVF-SAM\cite{evf} & 0.6290 & 0.7777 \\
         
         Robo-SAM (Ours) & \textbf{0.8620} & \textbf{0.9037} \\
         \bottomrule
    \end{tabular}
    \vspace{-2mm}
\end{table}

\begin{table*}[ht]
\centering
\vspace{2mm}
\caption{Comparison of Different Augmentation Methods on real robot performance, using both behavior score and success rate as metrics. All data are collected from a single scene, and the evaluation is conducted across 6 completely new scenes. See \autoref{fig:real-exp} (b) for scene visualization. }
\renewcommand{\arraystretch}{1.25}  
\begin{tabular}{lcccc|c}
\toprule
\rowcolor{gray!20} \textbf{} & \textbf{Fold Towel (Good Grasp)  $\uparrow$} & \textbf{Fold Towel (Finish)  $\uparrow$} & \textbf{Put Mouse (Good Grasp)  $\uparrow$} & \textbf{Put Mouse (Finish)  $\uparrow$} & \textbf{Average  $\uparrow$} \\ \midrule
\textbf{No aug}     & 0.36 / 40.6\%           & 0.29 / 15.6\%       & 0.15 / 6.0\%           & 0.07 / 0.0\%       & 0.20 / 15.6\% \\ 
\textbf{Inpainting} & 0.36 / 40.6\%           & 0.34 / 34.4\%       & 0.63 / 12.5\%          & 0.10 / 0.0\%       & 0.24 / 21.8\% \\ 
\textbf{Background} & 0.50 / 53.1\%           & 0.54 / 62.5\%       & 0.46 / 37.5\%          & 0.32 / 18.8\%      & 0.45 / 43.0\% \\ 
\textbf{ImageNet}   & 0.50 / 53.1\%           & 0.52 / 62.5\%       & 0.56 / 43.7\%          & 0.39 / 18.8\%      & 0.48 / 44.5\% \\ 
\textbf{Texture}    & 0.50 / 50.0\%           & 0.54 / 62.5\%       & 0.63 / 56.2\%          & 0.44 / 25.0\%      & 0.51 / 48.4\% \\ 
\textbf{\texttt{RoboEngine}}     & \textbf{0.56 / 56.2\%}           & \textbf{0.59 / 68.7\%}       & \textbf{0.79 / 75.0\%}          & \textbf{0.58 / 43.7\%}      & \textbf{0.62 / 60.9\%} \\ 
\bottomrule
\end{tabular}
\label{tab:real-robot}
\end{table*}

\subsection{Augmentation Baseline and Results}

We next compare our proposed \texttt{RoboEngine} augmentation with previous augmentation methods. For convenience, we have encapsulated all of these methods into our toolkit, allowing users to easily switch between them by adjusting a configuration parameter (``aug\_method''). Note that, for calibration-free and plug-and-play usage, all of these methods \textbf{depend on the \texttt{Robo-SAM} segmentation} model we proposed earlier. The baselines we compare with are as follows:

\begin{itemize}
    \item \textbf{Background}\cite{greenaug}: generates a scene image using a generative model from a scene description (which does not consider physics feasibility). The description is randomly selected from the description pool. 
    \item \textbf{Inpainting}\cite{semantic_aug, roboagent}: uses SAM-V2~\cite{sam2} to obtain masks for all objects, then inpaints 5 task-irrelevant objects.
    \item \textbf{ImageNet}\cite{rovi-aug}: randomly selects an image from the ImageNet\cite{imagenet} training set as the background.
    \item \textbf{Texture}\cite{greenaug}: randomly selects a texture as the background. 
    \item \textbf{No aug}\cite{data_scaling_law, dp}: no generative augmentation. 
\end{itemize}

For the ``Background'' method, we use Stable Diffusion V2.1\footnote{https://huggingface.co/stabilityai/stable-diffusion-2-1} as the generative model. For the ``Inpainting'' method, we use Stable Inpainting\footnote{https://huggingface.co/docs/diffusers/using-diffusers/inpaint}. For the ``Texture'' method, we select textures from the MIL-Dataset~\cite{mil_data} \footnote{https://huggingface.co/datasets/eugeneteoh/mil\_data}. A visualization of all methods is shown in \autoref{fig:real-exp} (a). It can be seen that \texttt{RoboEngine} is the only method that provides both high diversity and physics feasibility, aligning most closely with robot deployment environments.

\subsection{Real Robot Setting}

Next, we train real robot policy on these augmented method to evaluate the effectiveness of all augmentation methods.

\textbf{Robot and Policy Training.}\ \  We follow the hardware setup of the DROID platform~\cite{droid}, using a Franka Panda robot arm with a Robotiq gripper and absolute pose control, but only with one third-view RGB image stream. For policy training, we adopt the Diffusion Policy (DP)\cite{dp}, a state-of-the-art algorithm for robot learning. We make the following modifications: (1) Replace the original image encoder in DP with a DINOV2-Base\cite{dinov2} encoder to improve visual understanding~\cite{data_scaling_law, open-television}. (2) Set the observation horizon to 1 to better suit the current augmentation setup. We also set the action horizon to prediction horizon, as we find it performs better with this setting~\cite{rise}. (3) Train all tasks for 1000 epochs to ensure model convergence. For other hyperparameters, we follow the guidelines in~\cite{data_scaling_law}. Traditional image augmentation techniques, such as ColorJitter and RandomCrop, are applied across all methods during DP training~\cite{data_scaling_law}.

\textbf{Tasks and Metric.}\ \ We evaluate using two tasks: ``Fold Towel'' and ``Put Mouse on the Pad''. These tasks cover long-horizon (multi-stage), deformable (folding towel), and precise (grasping mouse) manipulation. Since sparse success rates (SR) may obscure performance differences, we also record the behavior score for each task. Detailed criteria for SR and the score can be found in Appendix~\ref{app: criteria}. For clarity, we normalize the behavior score by dividing the original score by the maximum possible score in \autoref{tab:real-robot}, resulting in a normalized maximum score of 1. Both tasks are multi-stage (grasping and manipulation stages), so we record SR and scores for both grasping and task completion.

\begin{figure*}[t]
    \centering
    \vspace{-2mm}
    \includegraphics[width=1.0\linewidth]{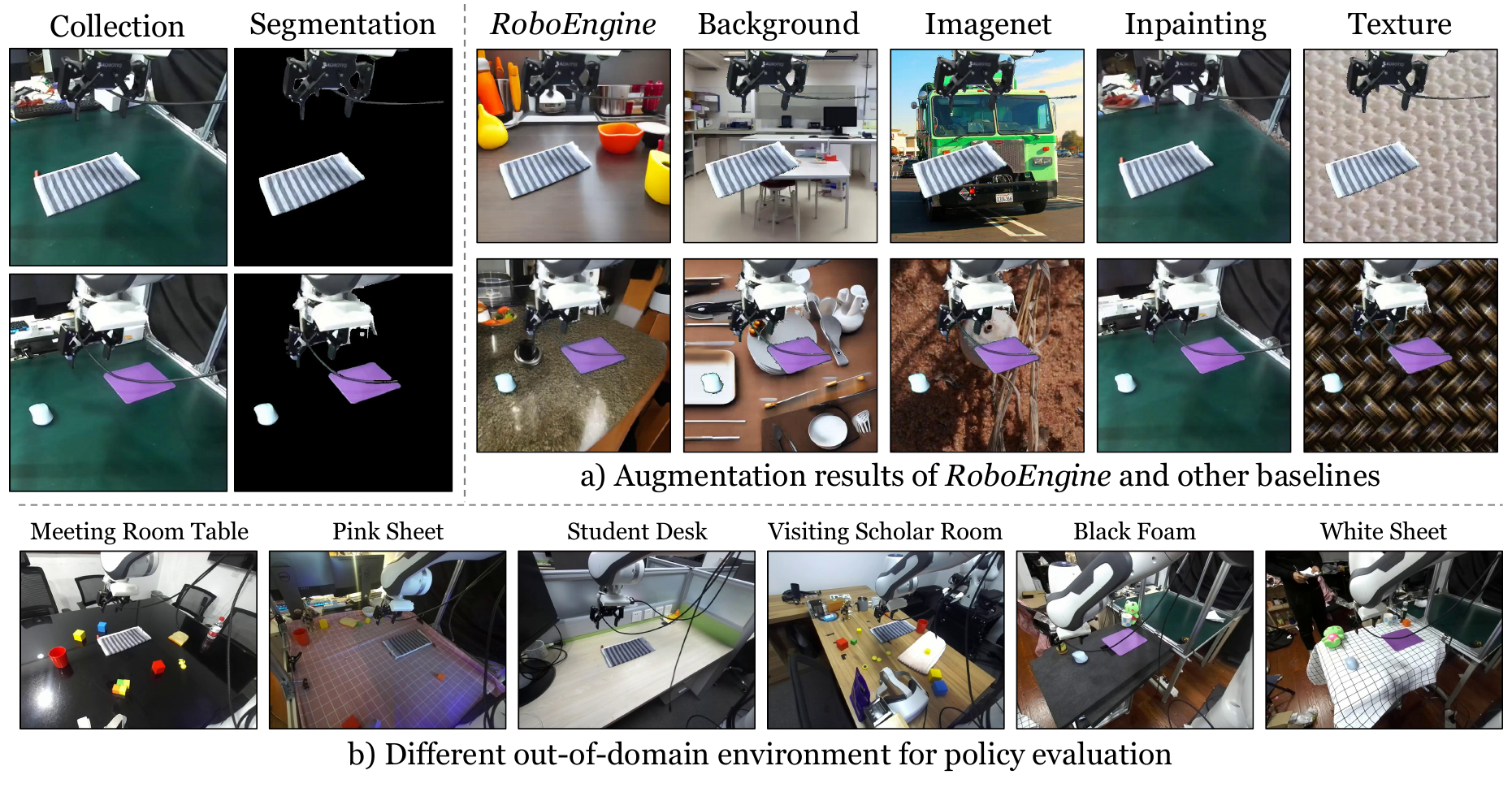}
    \vspace{-7mm}
    \caption{(a) Augmentation results using different methods. \texttt{RoboEngine} is the only method that simultaneously satisfies both physics constraints and high visual diversity. (b) Visualization of real robot evaluation environment. All scenes exhibit significant visual differences from the scene used for data collection. Video visualizations are available \href{https://roboengine.github.io/}{https://roboengine.github.io/}.}
    \vspace{-3mm}
    \label{fig:real-exp}
\end{figure*}

For the ``Fold Towel'' task, we collect 50 demonstrations, with towels placed in a 35cm × 35cm grid. For the ``Put Mouse'' task, we collect 100 demonstrations, with the mouse and pad placed in a 15cm × 15cm grid. By default, we apply each augmentation method to the demonstrations once, so the augmented dataset still contains 50 and 100 demos. Data is only collected in one scene, and we test the two tasks in 4 and 2 completely new scenes, respectively. The scene visualizations can be found in \autoref{fig:real-exp} (b). Note that our project focuses on solving visual generalization, not spatial generalization, so tables in all scenes share a similar height range of 73 to 76 cm.

\begin{table}[!t]
    \caption{Augmentation speed comparison across methods. The time (in seconds) required for augmenting a single frame is measured for each method.}
    \label{tab: speed}
    \centering
    \renewcommand{\arraystretch}{1.25}  
    \begin{tabular}{|c|c|c|c|c|}
    \hline
        \textbf{Inpainting} & \textbf{Background} & \textbf{Imagenet} & \textbf{Texture} & \textbf{\texttt{RoboEngine}} \\ \hline
        3.90 sec & 1.91 sec &  0.97 sec & 0.97 sec & 2.17 sec \\ \hline
    \end{tabular}
\end{table}

\subsection{Policy Evaluation Results}

We report the average score and success rate for all tasks in \autoref{tab:real-robot}, with per-scene results available in Appendix~\ref{app: raw_result}. Additionally, the time cost for all methods in the ``Fold Towel'' task is provided in \autoref{tab: speed} (with batch size set to 1 for measurement). Our method demonstrates a significant improvement, outperforming the best baselines by 20\%, and achieves a 210\% performance boost compared to the no-augmentation baseline.

For the ``Inpainting'' method~\cite{semantic_aug, roboagent}, we observe that performance heavily depends on the task-irrelevant object masks and is very sensitive to hyperparameters (e.g., the number of inpainting objects and their sizes). This often results in small visual changes, which are insufficient for robust visual generalization. Additionally, the time cost for this method is the highest, so we do not recommend it for practical use.

For the ``Background'', ``ImageNet'', and ``Texture'' methods, all show substantial performance gains compared to ``No aug'', with ``Texture'' outperforming the others slightly (consistent with findings in~\cite{greenaug}). Although the performance is lower than \texttt{RoboEngine}, both ``Texture'' and ``ImageNet'' are faster, as they do not require running a foundation generative model like Stable Diffusion~\cite{bg_diffusion}. We recommend using \texttt{RoboEngine} when computational resources are available for optimal performance, and opting for ``Texture'' or ``ImageNet'' when resources are limited.

\subsection{Performance Scaling Trend}

\begin{figure}[t]
    \centering
    \includegraphics[width=0.9\linewidth]{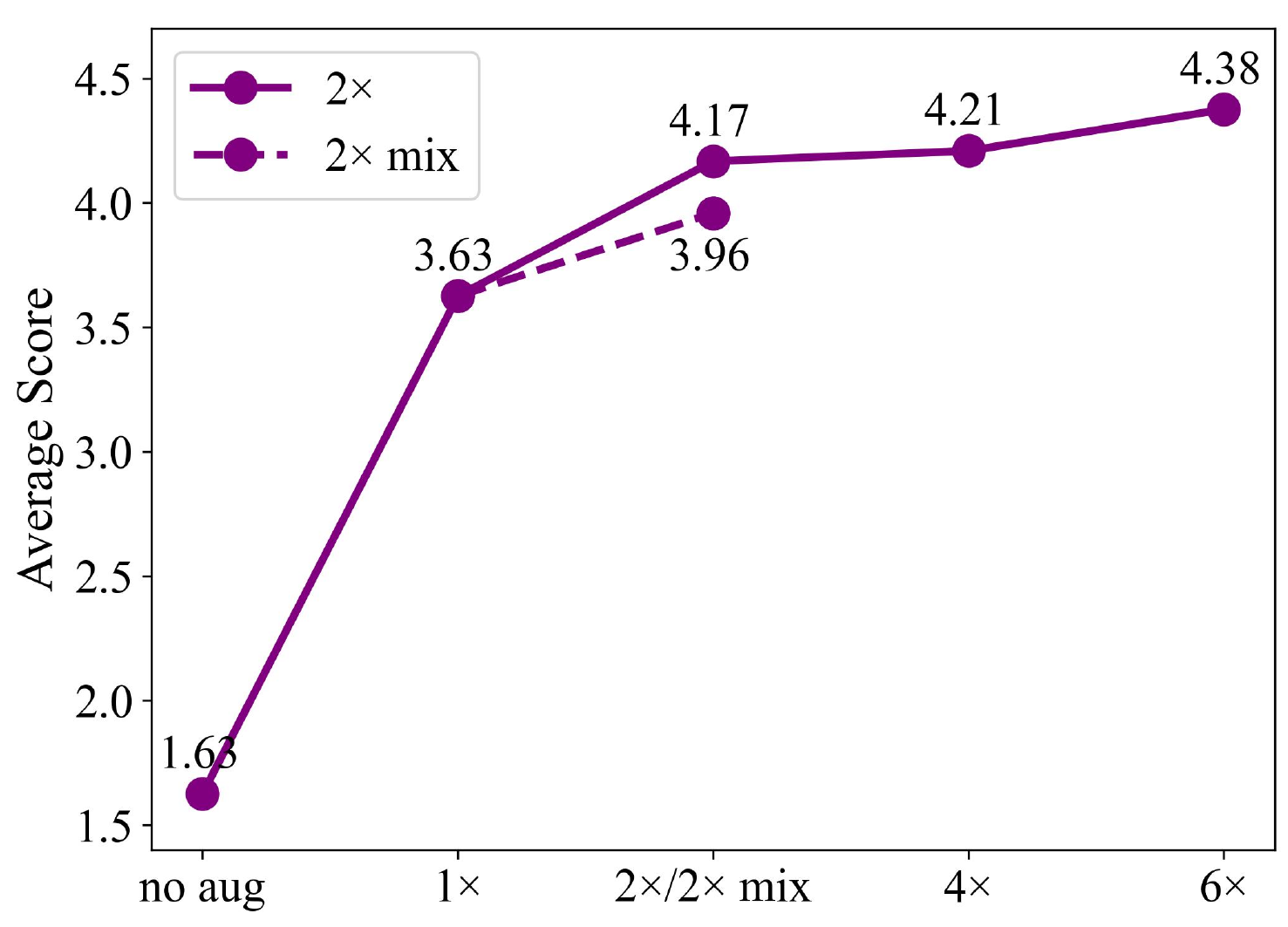}
    \caption{Performance scaling trend experiment results. We report the average behavior score (un-normalized) for the ``Fold Towel (Finish)'' task across 3 novel scenes. Raw results can be found in Appendix~\ref{app: scaling_exp}.}
    \label{fig:scaling_trend}
\end{figure}

We conduct an experiment to verify whether increasing the number of augmented demonstrations leads to better performance. For efficiency, we focus on the ``Fold Towel (Finish)'' task and evaluate the policy in 3 novel scenes (Meeting Room Table, Pink Sheet, Student Desk). We use our proposed \texttt{RoboEngine} augmentation method, which performs best among all methods (as shown in \autoref{tab:real-robot}). We represent the original augmented dataset as ``$1\times$'' (50 demos), and ``$N\times$'' represents datasets enlarged by a factor of $N$ (e.g., $50\times N$ demos). We also investigate whether mixing the original data with augmented data benefits policy training, using a ``$2\times \text{mix}$'' variant, which includes 50 original demos and 50 augmented demos. For a fair comparison, we adjust the number of training epochs for each variant to ensure the training steps are similar~\cite{data_scaling_law}. More details can be found in Appendix~\ref{app: scaling_exp}.

The results, shown in \autoref{fig:scaling_trend}, reveal two key findings: (1) Mixing the original dataset with augmented data does help to improve performance (``$2\times \text{mix}$'' vs. ``$1\times$''), but show no advantage to only using augmented data (``$2\times \text{mix}$'' vs. ``$2\times$''). (2) Increasing the number of augmented demonstrations does lead to performance improvements, but the rate of improvement gradually slows down and eventually reaches a bottleneck.

\section{CONCLUSION}

\textbf{Conclusion.}\ \ In this work, we introduce \texttt{RoboEngine}, the first plug-and-play visual robot data augmentation toolkit that enables physics- and task-aware robot scene background generation with no pre-requirements. By fine-tuning existing state-of-the-art models on our newly proposed \texttt{RoboSeg} dataset, we create the first high-quality, generalizable robot segmentation model, \texttt{Robo-SAM}, along with a new physics- and task-aware background generation Diffusion model. We encapsulate the entire pipeline into an out-of-the-box toolkit, allowing users to perform augmentation with just a few lines of code. Extensive real-robot experiments demonstrate the effectiveness of \texttt{RoboEngine} augmentation. We hope our work will help alleviate visual robustness issues in imitation learning and serve as a standard augmentation toolkit for the robot learning community.

\textbf{Limitation.}\ \ While \texttt{RoboEngine} provides significant convenience for visual robot data augmentation, it has some limitations. First, our method does not handle temporal consistency between frames, which could be addressed by recent video diffusion models~\cite{video_inp}. Second, we do not consider multi-view or 3D augmentation. This limitation may be overcome by using depth estimation techniques~\cite{video_depth_anything} to lift the generated background into 3D, followed by re-rendering.




\section*{APPENDIX}

\subsection{Details of Model Training} \label{app: model_training}

\textbf{Robo-SAM.}\ \ We choose EVF-SAM~\cite{evf} as our base model. EVF-SAM has two versions: the original version for instance segmentation and the ``multitask'' version for semantic segmentation\footnote{https://github.com/hustvl/EVF-SAM}. We find that the latter version significantly outperforms the former, therefore, we choose the ``multitask'' version as our base model. For fine-tuning, we use 3562 images as the training set and 92 images as the validation set. All other images from \texttt{RoboSeg} are used for model evaluation. We fine-tune the model for 30 epochs, with a learning rate of 1e-5 and a batch size of 32.

\textbf{BackGround-Diffusion.}\ \ We fine-tune BackGround-Diffusion~\cite{bg_diffusion} on the \texttt{RoboSeg} dataset for 100 epochs, with a learning rate of 5e-3 and a batch size of 32. The scene descriptions of the images are generated by GPT-4o~\cite{gpt-4o}. All images in \texttt{RoboSeg} are used for fine-tuning.

\subsection{Evaluation Criteria} \label{app: criteria}

The criteria for the behavior score are as follows:

\textbf{Fold Towel (Grasp Stage).}\\ 
0: The gripper does not contact the towel. \\
1: The gripper grasps the towel, but not at the left edge. \\
2: The gripper grasps at the left edge, but is outside a 7.5 cm range around the center of the edge. \\
3: The grasping is within the 7.5 cm perfect area of the left edge.

\textbf{Fold Towel (Fold Stage).}\\ 
0: Not folded. \\
1: The overlay is smaller than 1/3 of the maximum overlay. \\
2: The overlay is larger than 1/3 but smaller than 2/3. \\
3: The overlay is larger than 2/3 of the maximum overlay.

\textbf{Put Mouse (Grasp Stage).}\\ 
0: No contact with the mouse. \\
1: Contact with the mouse, but not grasped stably, or dropped during lifting. \\
2: The mouse is grasped stably, but has moved more than 3 cm on the table before grasping. \\
3: The mouse is grasped stably with small movement on the table.

\textbf{Put Mouse (Put Stage).}\\ 
0: The mouse is dropped during the put, and falls off the pad. \\
1: The mouse is dropped during the put, but falls on the pad. \\
2: The mouse is put on the pad stably, but outside the 10 cm × 10 cm center area. \\
3: The mouse is put in the center area stably.

For each stage, we consider it a success execution if the behavior score is greater than or equal to 2.

\begin{table*}[!t]
\centering
\vspace{2mm}
\caption{The raw evaluation results for all methods are presented, with the behavior score as the metric (un-normalized score). The success rate can be calculated based on the criteria outlined in Appendix~\ref{app: criteria}.}
\label{tab: raw_result}
\resizebox{\textwidth}{!}{
\begin{tabular}{l|rrrr|rrrr|rr|rr}
\rowcolor{gray!30}  
             & \multicolumn{4}{c}{Fold Towel (Grasp) $\uparrow$} & \multicolumn{4}{c}{Fold Towel (Finish) $\uparrow$} & \multicolumn{2}{c}{Put Mouse (Grasp) $\uparrow$} & \multicolumn{2}{c}{Put Mouse (Finish) $\uparrow$} \\
\toprule
             & Student Desk & Pink Sheet & Meeting Room & Scholar Room & Student Desk & Pink Sheet & Meeting Room & Scholar Room & Black Boam & White Sheet & Black Boam & White Sheet \\
\midrule
No aug       & 0.875        & 1.125      & 1.25         & 1.125        & 1.625        & 1.625      & 1.625         & 2            & 0.875      & 0           & 0.875      & 0           \\
Inpainting   & 0.625        & 1.625      & 0.875        & 1.25         & 1.375        & 3          & 1.75          & 2.125        & 1.25       & 0           & 1.25       & 0           \\
Background   & 1.875        & 1.25       & 1.5          & 1.375        & 4.125        & 2.875      & 3            & 2.875        & 1.625      & 1.125       & 2.375      & 1.5         \\
ImageNet     & 1.75         & 1.5        & 1.25         & 1.5          & 3.75         & 2.75       & 2.875         & 3.125        & 1          & 2.375       & 1.25       & 3.375       \\
Texture      & 1.875        & 1.5        & 1.5          & 1.125        & 4.125        & 3.5        & 3            & 2.25         & 1.375      & 2.375       & 2.125      & 3.125       \\
\texttt{RoboEngine} & 2           & 1.75       & 1.5          & 1.5          & 4.5          & 3.25       & 3.125         & 3.375        & 2.5        & 2.25        & 3.875      & 3.125       \\
\bottomrule
\end{tabular}
}
\vspace{-3mm}
\end{table*}

\subsection{Raw Evaluation Results} \label{app: raw_result}

For each task, we perform 8 evaluations per scene and compute the average behavior score. The raw evaluation results are provided in \autoref{tab: raw_result}.

\begin{table}[!t]
\centering
\caption{Raw results of the scaling trend experiment, conducted on the 'Fold Towel (Finish)' task, with the behavior score (un-normalized) as the metric.}
\label{tab: scaling_raw_result}
\begin{tabular}{lrrr}

\rowcolor{gray!30} 
       & \multicolumn{1}{c}{Pink Table $\uparrow$} & \multicolumn{1}{c}{Meeting Room $\uparrow$} & \multicolumn{1}{c}{Student Table $\uparrow$} \\
\toprule
$2\times \text{mix}$  & 4.375                          & 2.875                            & 4.625                                                   \\
No aug & 1.625                          & 1.625                              & 1.625                                                   \\
$1\times$     & 3.25                           & 3.125                            & 4.5                                                \\
$2\times$     & 4.5                            & 2.875                            & 5.125                                                  \\
$4\times$     & 5                              & 3                                & 4.625                                               \\
$6\times$     & 4.625                          & 3.375                            & 5.125                                                   \\
\bottomrule
\end{tabular}
\end{table}

\subsection{Details of Scaling Trend Experiment} \label{app: scaling_exp}

We train all variants with the following settings: (1) ``$1 \times$'' for 1000 epochs, (2) ``$2 \times$'' and ``$2 \times \text{mix}$'' for 700 epochs, (3) ``$4 \times$'' for 400 epochs, and (4) ``$6 \times$'' for 300 epochs. For resource efficiency, we only conduct experiments on the ``Fold Towel (Finish)'' task. For each policy and scene combination, we perform 8 rollouts and record the average behavior score. The raw results can be found in \autoref{tab: scaling_raw_result}.


\section*{ACKNOWLEDGMENT}

This work is supported by the National Key R\&D Program of China (2022ZD0161700), National Natural Science Foundation of China (62176135), Shanghai Qi Zhi Institute Innovation Program SQZ202306 and the Tsinghua University Dushi Program, the grant of National Natural Science Foundation of China (NSFC) 12201341.


\bibliographystyle{IEEEtran}
\bibliography{references}

\end{document}